\documentclass[10pt,twocolumn,letterpaper]{article}

\usepackage{wacv}
\usepackage{times}
\usepackage{epsfig}
\usepackage{graphicx}
\usepackage{amsmath}
\usepackage{amssymb}
\usepackage{multirow}
\usepackage{multicol}
\usepackage[T1]{fontenc}
\usepackage[utf8]{inputenc}
\usepackage{booktabs}
\usepackage{authblk}
\usepackage{float}

\def\wacvPaperID{766} 

\wacvfinalcopy
\ifwacvfinal
\def\assignedStartPage{1} 
\fi

\ifwacvfinal
\usepackage[breaklinks=true,bookmarks=false]{hyperref}
\fi

\ifwacvfinal
\else
\fi

\begin{document}

\title{RarePlanes: Synthetic Data Takes Flight}

\renewcommand*{\Authfont}{\mdseries}
\author[1]{Jacob Shermeyer}
\author[2]{Thomas Hossler}
\author[1]{Adam Van Etten}
\author[1]{Daniel Hogan}
\author[1]{Ryan Lewis}
\author[2]{Daeil Kim}
\affil[1]{IQT - CosmiQ Works, [jshermeyer, avanetten, dhogan, rlewis]@iqt.org}
\affil[2]{AI.Reverie, [thomas.hossler, daeil]@aireverie.com}

\maketitle

\begin{abstract}
RarePlanes is a unique open-source machine learning dataset that incorporates both real and synthetically generated satellite imagery. The RarePlanes dataset specifically focuses on the value of synthetic data to aid computer vision algorithms in their ability to automatically detect aircraft and their attributes in satellite imagery. Although other synthetic/real combination datasets exist, RarePlanes is the largest openly-available very-high resolution dataset built to test the value of synthetic data from an overhead perspective. Previous research has shown that synthetic data can reduce the amount of real training data needed and potentially improve performance for many tasks in the computer vision domain. The real portion of the dataset consists of $253$ Maxar WorldView-3 satellite scenes spanning $112$ locations  and $2,142 \, {\rm km}^2$ with $14,700$ hand-annotated aircraft. The accompanying synthetic dataset is generated via AI.Reverie's simulation platform and features $50,000$ synthetic satellite images simulating a total area of $9331.2\, \rm{km}^2$  with $\sim630,000$ aircraft annotations. Both the real and synthetically generated aircraft feature $10$ fine grain attributes including: aircraft length, wingspan, wing-shape, wing-position, wingspan class, propulsion, number of engines, number of vertical-stabilizers, presence of canards, and aircraft role. Finally, we conduct extensive experiments to evaluate the real and synthetic datasets and compare performances. By doing so, we show the value of synthetic data for the task of detecting and classifying aircraft from an overhead perspective.
\end{abstract}

\section{Introduction}

Over the last decade, computer vision research and the development of new algorithms has been driven largely by permissively licensed open datasets. Datasets such as ImageNet \cite{deng2009imagenet}, MSCOCO \cite{lin2014microsoft}, and PASCALVOC \cite{everingham2010pascal} (among others) remain critical drivers for advancement.  Convolutional neural networks (CNNs), currently the leading class of algorithms for most vision tasks \cite{Class_rev,OD_rev}, require a large amount of annotated observations. However, the development of such datasets is often manually intensive, time-consuming, and costly to create. An alternative approach to manually annotating training data is to create computer generated images and annotations (referred to as synthetic data). After creating realistic 3D environments, one can then generate thousands of images at virtually no cost.  Such data has been shown to be effective for augmenting and replacing real data, thus reducing the burden of dataset curation. Synthetic datasets continue to be developed and have been notably helpful in various domains including: autonomous driving \cite{playforbenchmarks, playfordata, synthia, vKITTI}, optical flow \cite{flow_synth,playforbenchmarks,free}, facial recognition \cite{kortylewski2018faces, 3dmorph, kortylewski2018synthetic}, amodal analysis \cite{SAIL-VOS,DYCE} and domain adaptation \cite{DA_review,MUNIT,cycada,ADVENT} (see Section \ref{sec:synth_data} for further detail).

\begin{figure*}[!htb]
\centering
  \includegraphics[width=\textwidth]{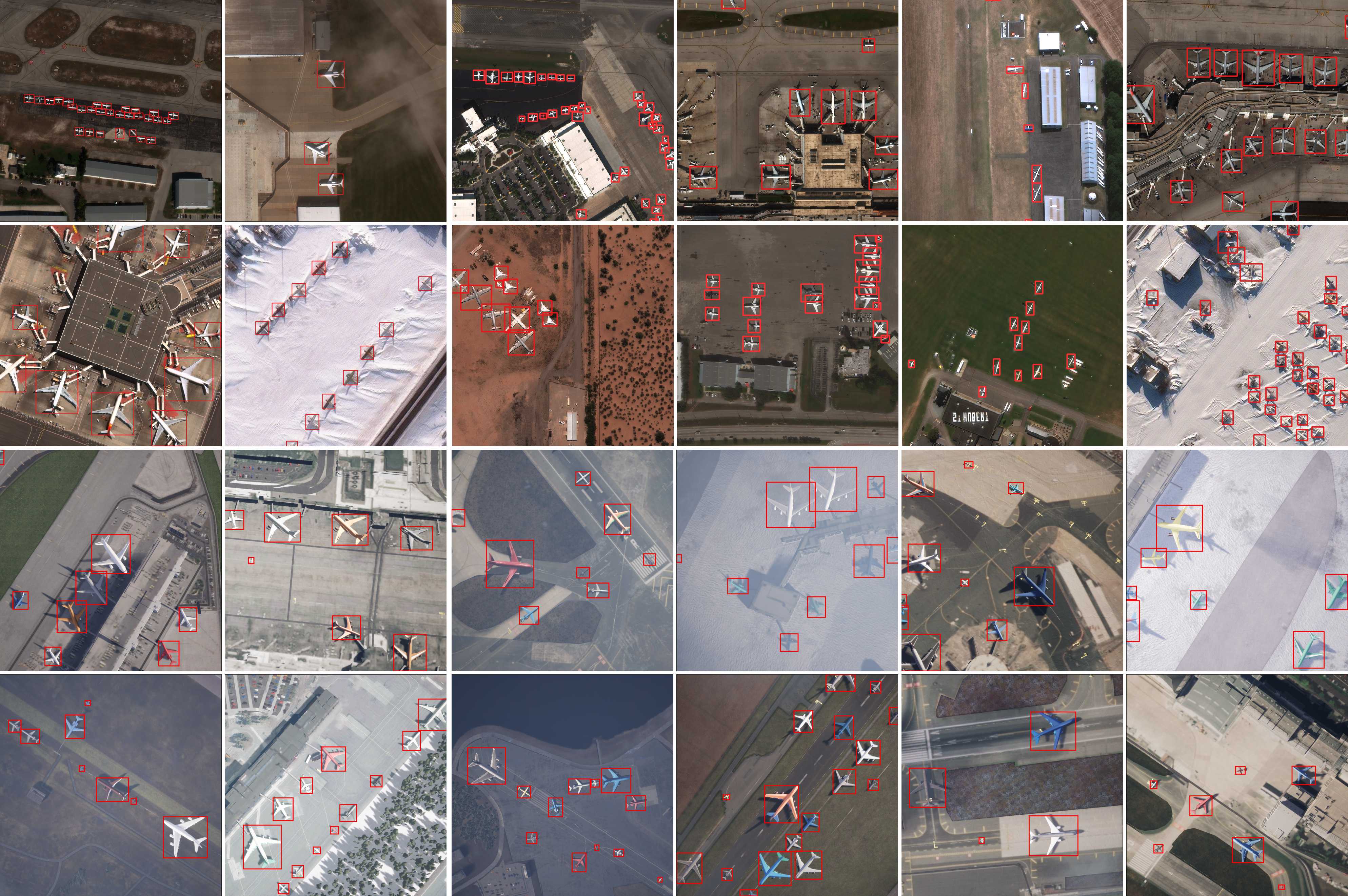}\caption{\textbf{Example of the real and synthetic datasets present in RarePlanes.} The top two rows feature the real Maxar WorldView-3 satellite imagery  and the bottom two rows show the synthetic data. The dataset features variable weather conditions, biomes, and ground surface types. }
  \label{Real_Synthetic_Data}
    \vspace{-2mm}
\end{figure*}

Although synthetic datasets continue to become more prevalent, no expansive permissively licensed synthetic datasets exist in the context of overhead observation. Overhead imagery presents unique challenges for computer vision models such as: the detection of small visually-heterogeneous objects, varying look angles or lighting conditions, and different geographies that can feature distinct seasonal variability. As such, creating synthetic datasets from an overhead perspective is a significant challenge and simulators must attempt to closely mimic the complexities of a spaceborne or aerial sensor as well as the Earth's ever-changing conditions.  For example, to create a large and heterogeneous synthetic dataset, one must account for each sensors varying spatial resolution, changes in sensor look angle, the time of day of collection, shadowing, and changes in illumination due to the sun's location relative to the sensor.  Furthermore, the simulator must be able to account for other factors such as the ground appearance due to seasonal change, weather conditions, and varying geographies or biomes.

 While synthetic datasets certainly have the potential to be beneficial, they require a paired real dataset with shared features to baseline performance and quantitatively test value.  However, few permissively licensed overhead datasets \cite{etten2018spacenet,weir2019spacenet,shermeyer2020spacenet} exist that focus on detection or segmentation tasks and feature very-high resolution real imagery from an overhead perspective. Overhead datasets remain one of the best avenues for developing new computer vision methods that can adapt to limited sensor resolution, variable look angles, and locate tightly grouped, cluttered objects. Such methods can extend beyond the overhead space and be helpful in other domains such as face-id, autonomous driving, and surveillance.

To address the limitations described above, we introduce the RarePlanes dataset. This dataset focuses on the detection of aircraft and their fine-grain attributes from an overhead perspective.  It consists of both an expansive synthetic and real dataset.  We use AI.Reverie's novel platform which incorporates unreal engine\cite{unreal} to develop realistic synthetic data based off of real world airports. The platform ingests real world metadata and overhead images to procedurally generate 3D environments of real world locations. The weather, time of collection, sunlight intensity, look angle, biome, and distribution of aircraft model are among the multiple parameters that the simulator can modify to create diverse and heterogeneous data.  The synthetic portion of RarePlanes consists of $50,000$ images simulating a total area of $9331.2\, \rm{km}^2$ and  $\sim630,000$ annotations.  The real portion consists of $253$ Maxar WorldView-3 satellite images spanning $112$ locations in $22$ countries and $2,142\text{km}^2$ with $\sim14,700$ hand annotated aircraft.  Examples of the synthetic and real images are shown in Figure \ref{Real_Synthetic_Data}.

RarePlanes also provides fine-grain labels with $10$ distinct aircraft attributes and $33$ different sub-attribute choices labeled for each aircraft.  These include: aircraft length, wingspan, wing-shape, wing-position, Federal Aviation Administration (FAA) wingspan class \cite{gudmundsson2013general}, propulsion, number of engines, number of vertical-stabilizers, canards, and aircraft type or role.  Although several other overhead detection datasets exist \cite{etten2018spacenet,weir2019spacenet,shermeyer2020spacenet,Lam:2018,xbd,isaid}, no others have multiple fine-grain attributes that detail specific object features. Such fine-grain attributes have been particularly helpful for zero-shot learning applications \cite{ZSL} and enable end users to create diverse custom classes.  Using these combined attributes, anywhere from 1 to 110 classes can be created for individual research purposes. The dataset is available for free download through Amazon Web Services' Open Data Program, with download instructions available at \url{https://www.cosmiqworks.org/RarePlanes}.

\begin{table*}[!ht]
\centering
\begin{tabular}{lrrrrr}
\hline
\multirow{2}{*}{\textbf{Dataset}} &
  \multicolumn{1}{c}{\multirow{2}{*}{\textbf{Gigapixels}}} &
  \multicolumn{1}{c}{\multirow{2}{*}{\textbf{Classes}}} &
  \multicolumn{1}{c}{\multirow{2}{*}{\textbf{Attributes}}} &
  \multicolumn{2}{c}{\textbf{Labels}} \\ \cline{5-6} 
 &
  \multicolumn{1}{c}{} &
  \multicolumn{1}{c}{} &
  \multicolumn{1}{c}{} &
  \multicolumn{1}{c}{\textbf{Real}} &
  \multicolumn{1}{c}{\textbf{Synthetic}} \\ \hline
SpaceNet \cite{etten2018spacenet,weir2019spacenet,shermeyer2020spacenet}& 100.1   & 1 to 8     & 1 & 859,982  & 0       \\
xBD \cite{xbd}                                              & 9.8    & 1 to 4     & 1 & 850,736  & 0       \\
xView \cite{Lam:2018}                                       & 56.0   & 60    & 0 & 1,000,000 & 0       \\
iSAID \cite{isaid}                                            & 44.9   & 15    & 0 & 655,451  & 0       \\
FMOW \cite{fmow}                                            & 1084.0 & 63    & 0 & 132716  & 0       \\ 
\hline
Cityscapes \cite{Cityscapes} + GTA \cite{playforbenchmarks} & 537.5  & 30/19 & 0 & 210,179  & 510,4434 \\
COCOA \cite{COCOA} + SAIL-VOS \cite{SAIL-VOS}               & 115.7  & -/163 & 0 & 46,314   & 1,896,296 \\ \hline
Animals with Attributes 2 \cite{Animals_w_attributes2}                          & 24.7  & 50 & 85 & 37,322  & 0 \\
CompCars \cite{CompCars}                                   & 86.1  & 1,716 & 13 & 136,726   & 0 \\
\hline
\hline
\textbf{RarePlanes (Ours)}                  &
187.1 & 1 to 110 & 10 & 14,707 & 629,551 \\ \hline
\end{tabular}
\caption{\textbf{Comparison with other synthetic, attribute and overhead imagery datasets.} Our dataset has a similar scale as modern computer vision datasets and provides both a real and synthetic component.  For SpaceNet (Buildings + Road Speed), xBD (Building Damage Scale), and RarePlanes we report the range of possible customizable classes that end-users can create using varieties of the dataset attributes.} \vspace{3mm}
\label{datasets}
\end{table*}

\subsection*{Contributions}
\begin{itemize}
\item An expansive real and synthetic overhead computer vision dataset focused on the detection of aircraft and their features.
\item Annotations with fine-grain attributions that enable various CV tasks such as: detection, instance segmentation, or zero-shot learning. 
\item Extensive experiments to evaluate the real and synthetic datasets and compare performances. By doing so, we show the value of synthetic images for the task of detecting and classifying aircraft from an overhead perspective. 
\end{itemize}

\section{Related Work}

RarePlanes sits at the intersection of three distinct computer vision dataset domains: synthetic datasets, geospatial datasets, and fine-grain attribution datasets. These three domains are cornerstones around which computer vision research has continued to rapidly advance and grow.   We summarize the key characteristics of modern synthetic, geospatial, and attribute datasets in Table \ref{datasets} and compare them to the RarePlanes dataset.

\subsection{Synthetic Datasets}\label{sec:synth_data}

Synthetic data has become prevalent across many computer vision domains and has shown value as a replacement for real data or to augment existing training datasets \cite{playfordata,kortylewski2018synthetic, synthia,alhaija_instance,moreobj}.  Many synthetic datasets focus on the autonomous driving domain; including the Synthia \cite{synthia}, GTA \cite{playforbenchmarks, playfordata}, and vKITTI \cite{vKITTI} datasets.  These synthetic datasets are often paired with real-world data such as Cityscapes \cite{Cityscapes}, CamVid \cite{CamVid}, or KITTI \cite{KITTI} to benchmark the value of synthetic data. Other notable synthetic datasets such as SUNCG \cite{SUNCG} or Matterport3D \cite{Matterport3D} focus on indoor scenes and include RGB-D data for depth estimation. Moreover, other datasets focus on addressing challenging occlusion (amodal) problems such as the expansive SAIL-VOS \cite{SAIL-VOS} and DYCE \cite{DYCE}.  Finally, the Synthinel-1 \cite{synthinel2020} Dataset is the only other dataset that bridges the synthetic/satellite domain.  It features synthetic data from an overhead perspective with binary pixel masks of building footprints.  Overall, combined synthetic and real datasets, similar to RarePlanes, have been helpful with several different tasks including: enhancing object detection \cite{obj_domain,deep_obj,moreobj}, semantic segmentation \cite{synthia, semantic_handa,more_semantic}, or instance segmentation performance \cite{instance_ward,alhaija_instance}. Furthermore, such datasets continue to inspire new domain adaptation (DA) techniques \cite{DA_review,MUNIT,cycada,ADVENT,obj_domain}. Such DA techniques could be particularly valuable for overhead applications as there remains a dearth of openly available training data and models trained on one location often do not generalize well to new areas. 



\begin{figure}[!htb]
  \includegraphics[width=\columnwidth]{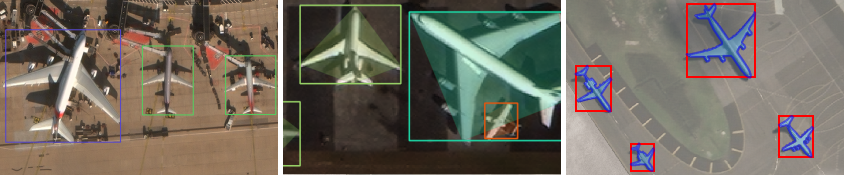}\caption{\textbf{Three annotation styles within RarePlanes.} The dataset features three annotation styles including: 'Bounding Box', 'Diamond Polygon', and 'Full Instance Segmentation' (synthetic only).}
  \label{Annotations}
\end{figure}

\subsection{Geospatial Datasets}

Geospatial and very-high resolution remote sensing datasets have continued to draw increased interest due to their relevancy to many computer vision challenges.  Such datasets contain lower resolution images with tiny, closely grouped objects with varying aspect ratios, arbitrary orientations and high annotation density. The lessons learned from such datasets continue to inspire new computer vision approaches related to detection \cite{more_OD,moreOD_2,yolt}, segmentation \cite{ternausnet}, super-resolution \cite{SR1,SR2}, and even bridges to natural language processing \cite{NLP}. Some notable datasets include SpaceNet \cite{etten2018spacenet,weir2019spacenet,shermeyer2020spacenet} and xBD \cite{xbd}, which focus on foundational mapping and instance/semantic segmentation for problems such as building footprint and road network extraction or building damage assessment. Others such as xView \cite{Lam:2018}, A large-scale dataset for object detection in aerial images (DOTA) \cite{dota} and A Large-scale Dataset for Instance Segmentation in Aerial Images (iSAID) \cite{isaid} focus on overhead object detection or instance segmentation, featuring multiple classes of different object types. The Functional Map of the World (FMOW) \cite{fmow} dataset centers on the task of classification of smaller image chips from an overhead perspective. Other datasets such as CORE3D \cite{core3d}, MVS \cite{bosch2016multiple}, the ISPRS benchmark \cite{isprs_3d} and SatStereo \cite{SatStereo} target height extraction and 3D mapping from space.   RarePlanes builds upon these existing datasets and contributes both synthetic and real data. Furthermore, RarePlanes adds 10 unique object attributes, which enable customizable classes, as well as three annotation styles per object (Bounding Box, Diamond Polygon, and Full-Instance (Synthetic Only)).

\subsection{Fine-Grain Attribute Datasets}

Many datasets focus on identifying general objects in imagery, however, several others take an alternative approach and label unique attributes of each object. As previously stated, RarePlanes features 10 attributes and 33 sub-attributes.  Such attribution has been particularly valuable for constructing new zero-shot learning methods and algorithms \cite{ZSL}. The Comprehensive Cars \cite{CompCars} dataset is similar to RarePlanes and features attribute labels of 5 car attributes and 8 car-parts, as well as different look angles of vehicles.  Several other similar datasets \cite{Animals_w_attributes2,aPY, Birds, SUN,largescale_attribute} feature multiple classes with extensive ranges in attributes; most of which are geared toward zero-shot learning research. 

\section{The RarePlanes Dataset and Statistics}

\subsection{Annotations, Features, and Attributes}

The RarePlanes dataset contains 14,707 real and 629,551 synthetic annotations of aircraft.  Each aircraft is labeled in a diamond style with the nose, left-wing, tail, and right-wing being labeled in successive order (Figure \ref{Annotations}). This annotation style has the advantage of being: simplistic, easily reproducible, convertible to a bounding box, and ensures that aircraft are consistently annotated (other hand-annotated formats can often lead to imprecise labeling). Furthermore, this annotation style enables the calculation of aircraft length and wingspan by measuring between the first annotation node to the third and from the second to the fourth. We employ a professional labeling service to produce high-quality annotations for the real portion of the dataset. Two rounds of quality control are included in the process, a first one by the professional service and a second by the authors.

\begin{figure}[H]
\centering
  \includegraphics[width=0.92\columnwidth]{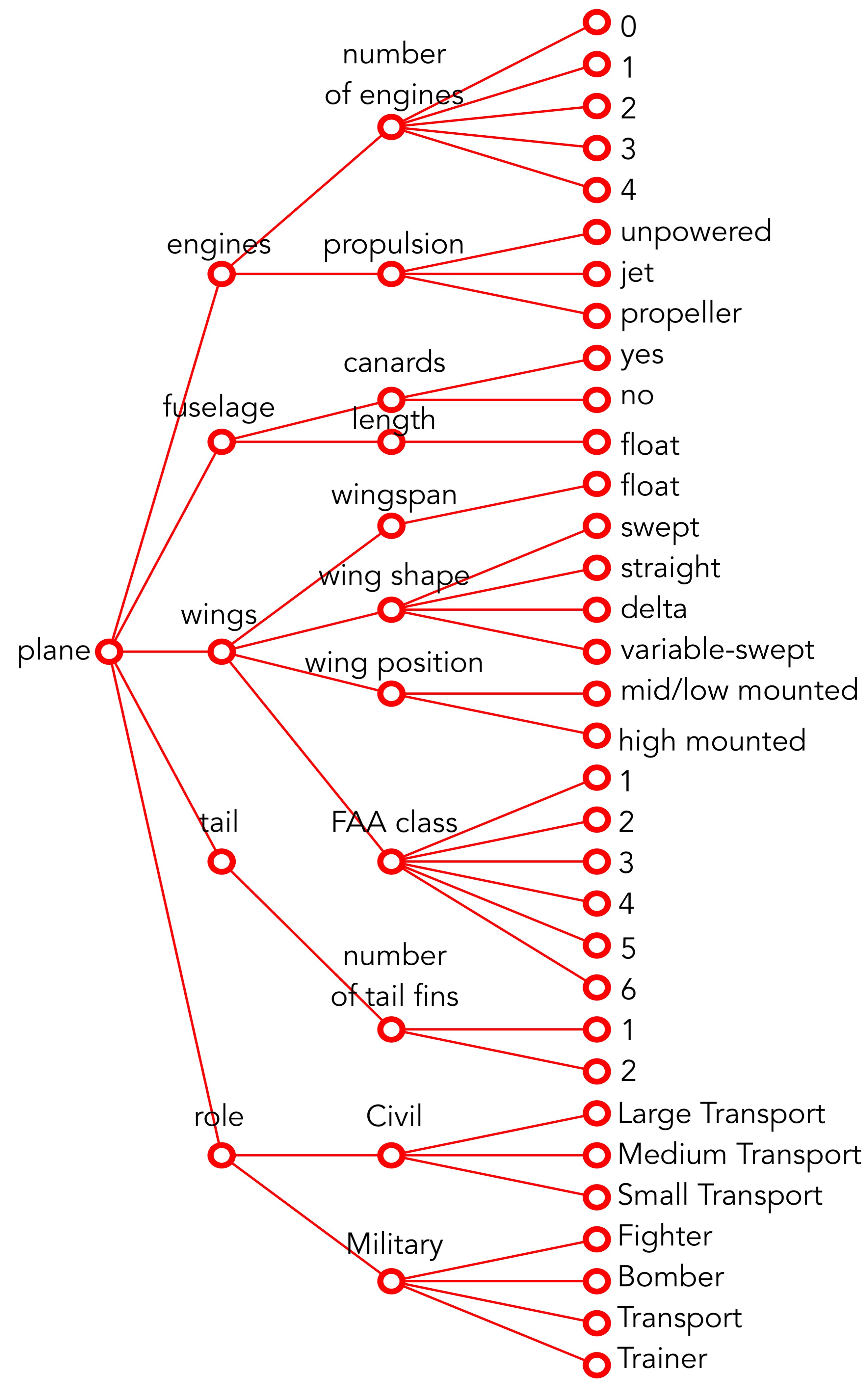}\caption{\textbf{The $5$ features, $10$ attributes, and $33$ sub-attributes contained in the RarePlanes dataset.} The dataset and associated codebase enables users to create custom classes using groupings of these attributes.}
  \label{Tree}
\end{figure}

After each aircraft is annotated in the diamond format, an expert geospatial team labels aircraft features.  The features include attributes of aircraft \textbf{wings}, \textbf{engines}, \textbf{fuselage}, \textbf{tail}, and \textbf{role} (Figure \ref{Tree}).   We ultimately chose these attributes as they were visually distinctive from an overhead perspective and have been shown to be helpful in aiding to visually identifying the type or make of aircraft \cite{janes}.

\begin{itemize}
    \item \textit{\textbf{Wings:}} We label aircraft \textbf{Wing Shape:} (`straight', `swept', `delta', and `variable-swept'), \textbf{Wing Position:}  (`high mounted' and `mid/low mounted'), \textbf{Wingspan in Meters:} (`float'), and the \textbf{FAA Aircraft Design Group Wingspan Class:} \cite{gudmundsson2013general} (`1' to `6') which determines which airports can accommodate different sized aircraft.  Examples of wing-shape and position can be seen in figure \ref{Wings}. \begin{figure} [H]
      \includegraphics[width=\columnwidth]{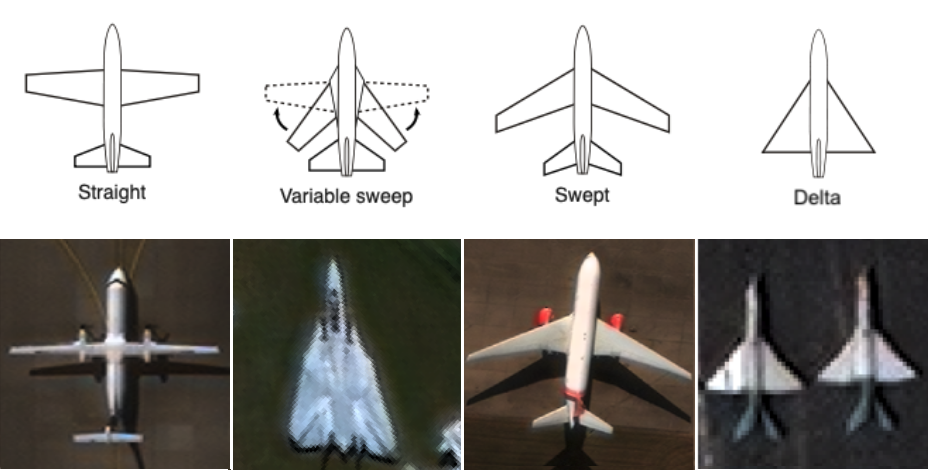}\caption{\textbf{Wing Shapes \cite{wiki:User:Steelpillow/Aircraft} present in the RarePlanes dataset.} Note that the two left-most aircraft feature `high-mounted' wings, with the two right-most aircraft featuring `mid/low mounted' wings.}
      \label{Wings}
    \end{figure}
    \item \textit{\textbf{Fuselage:}} We label aircraft \textbf{Length in Meters:} (`float') and if the plane has \textbf{Canards:} (`yes' or `no').  Canards are small fore-wings that are added to planes to increase maneuverability or reduce the load/airflow on the main wing.    
    \item \textit{\textbf{Engines:}} we label the \textbf{Number of Engines:} (`0' to `4') and the \textbf{Type of Propulsion:} (`unpowered', `jet', `propeller').   \begin{figure}[H]
\begin{center}
  \includegraphics[width=0.24\columnwidth]{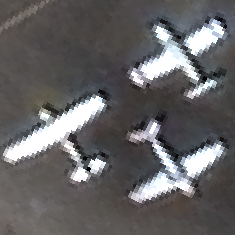}
  \includegraphics[width=0.24\columnwidth]{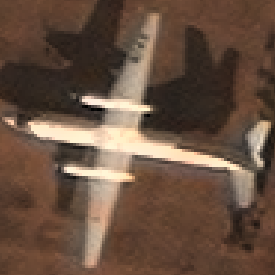}
  \includegraphics[width=0.24\columnwidth]{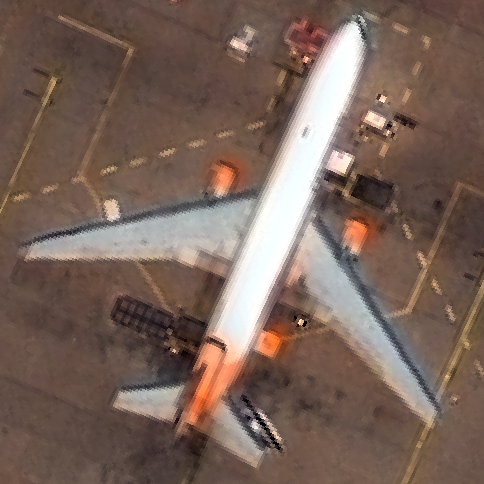}
  \includegraphics[width=0.24\columnwidth]{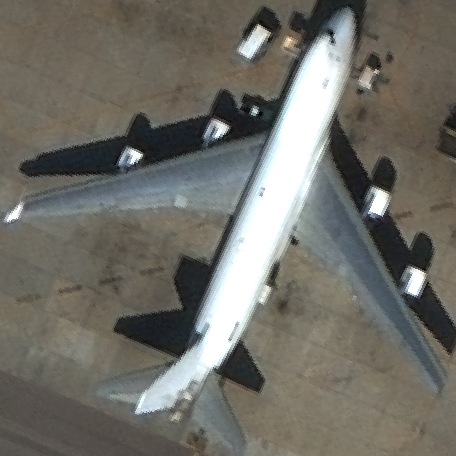}
  \caption{\textbf{Engine Types. From left to right:} A single engine propeller aircraft, a two engine propeller aircraft, a three engine jet aircraft, and a four engine jet aircraft. }
  \label{engines}
\end{center}
\end{figure}

     \item \textit{\textbf{Tail:}} We label the \textbf{Number of Vertical Stabilizers:} (`1' or `2') or tail fins that each aircraft possesses.
    \begin{figure} [H] \begin{center}
  \includegraphics[width=0.75\columnwidth]{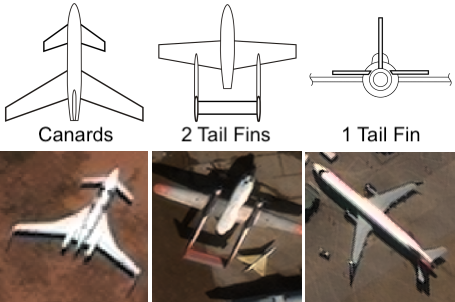}\caption{\textbf{Examples of fuselage and tail attributes including canards and number of vertical stabilizers \cite{wiki:User:Steelpillow/Aircraft} present in the RarePlanes dataset.} }
  \label{CanardsTailFins}
  \end{center}
\end{figure}
    \item \textit{\textbf{Role:}} After labeling each attribute, we then use these attributes to classify the \textbf{Role or Type:} of an aircraft into seven unique classes.  These include: 'Civil Transport/Utility' (`Small', `Medium, and `Large' based upon wingspan), `Military Transport/Utility/AWAC', `Military Bomber', `Military Fighter/Interceptor/Attack', and `Military Trainer'. Further detail on role definitions and can be found in the RarePlanes User Guide, hosted on AWS with the dataset.  \begin{figure}[H]
\begin{center}
  \includegraphics[width=0.25\columnwidth]{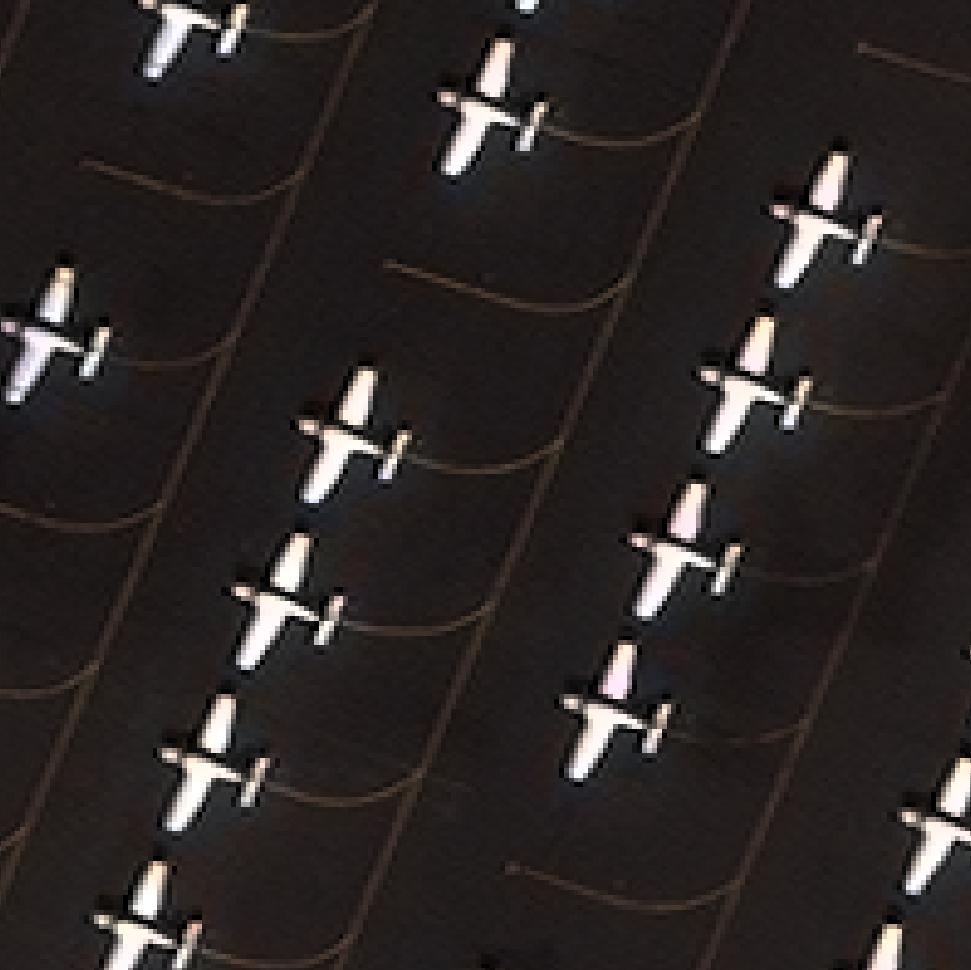}
  \includegraphics[width=0.25\columnwidth]{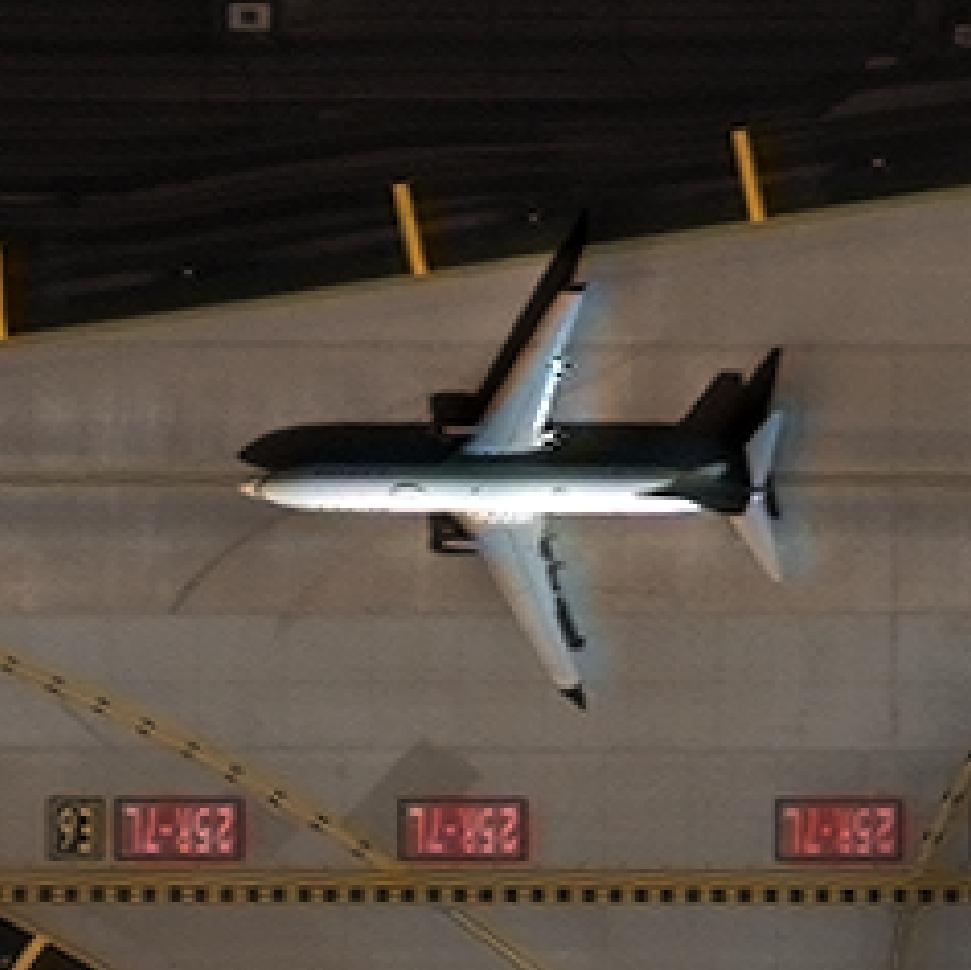}
  \includegraphics[width=0.25\columnwidth]{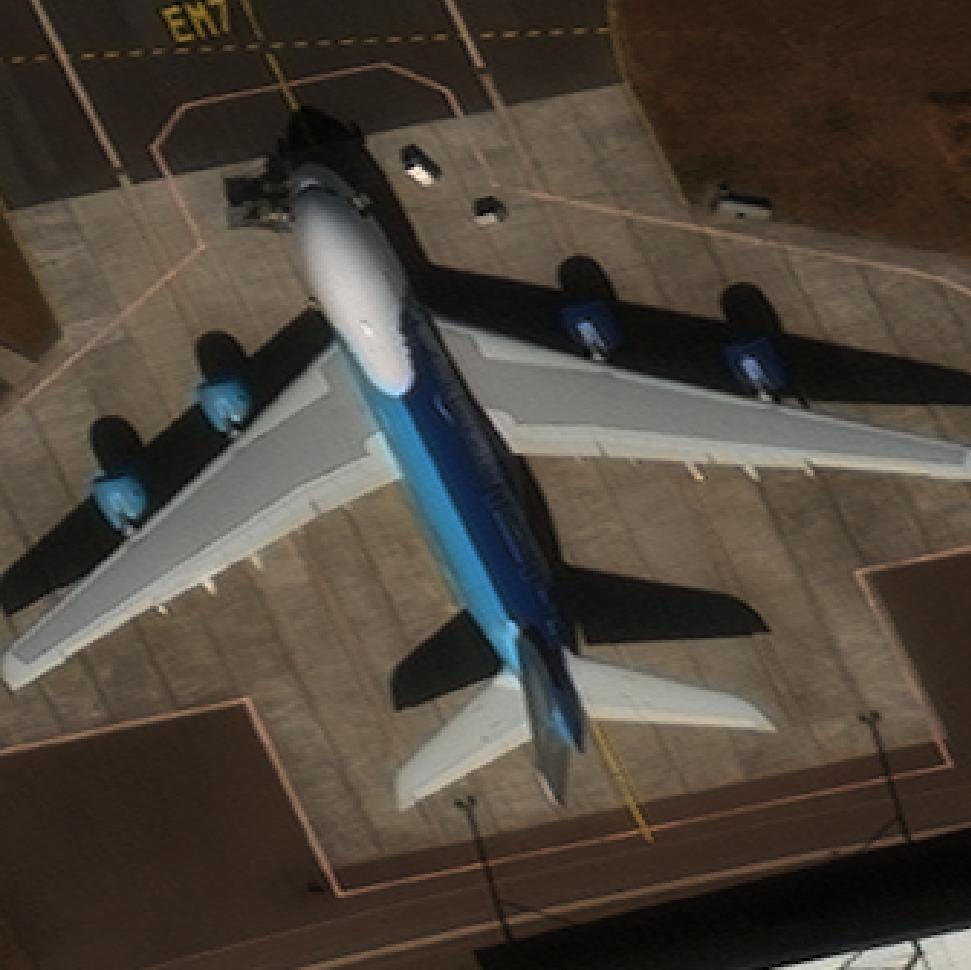}
  \includegraphics[width=0.25\columnwidth]{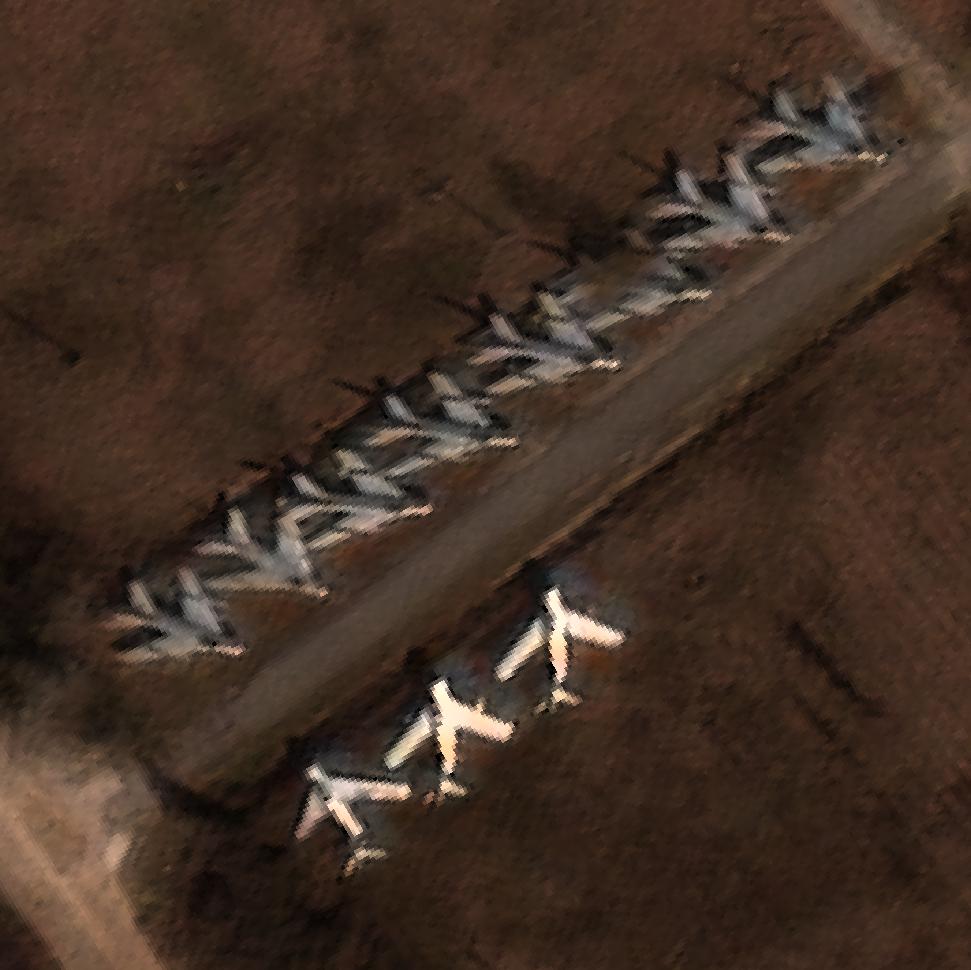}
  \includegraphics[width=0.25\columnwidth]{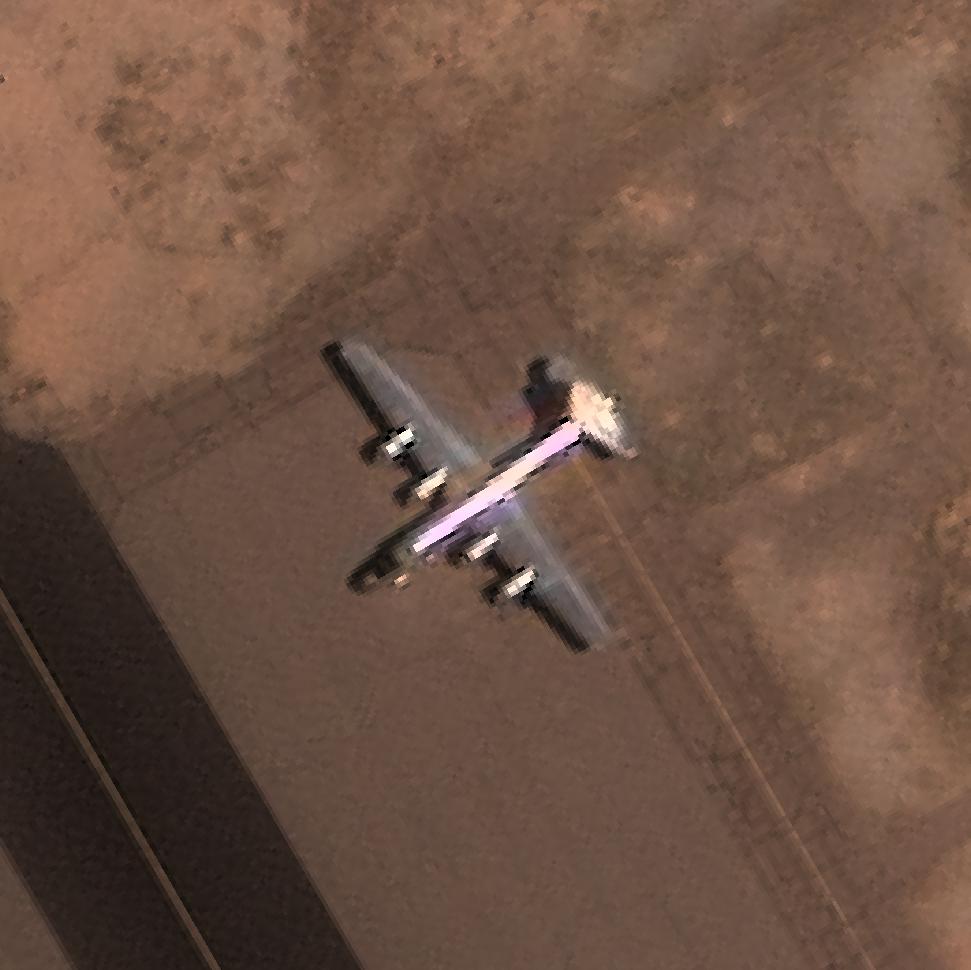}
  \includegraphics[width=0.25\columnwidth]{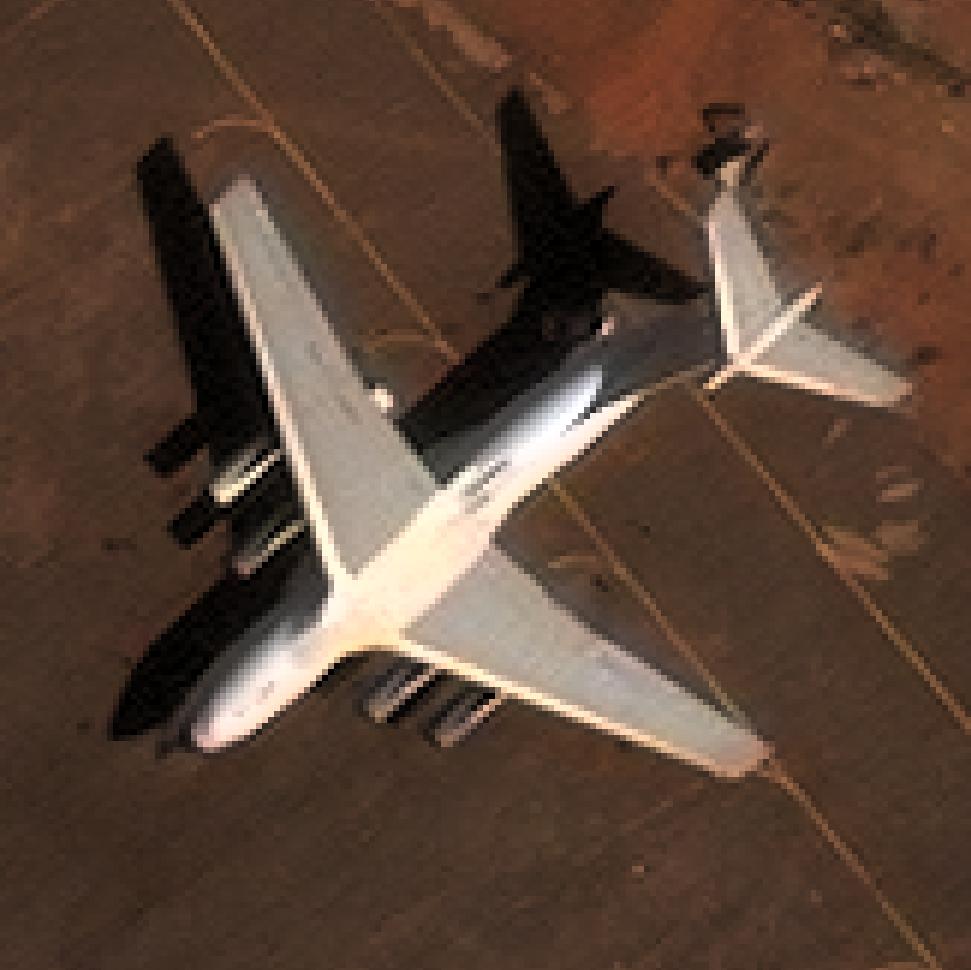}
  \caption{\textbf{Aircraft by role at 1:150 scale. Note the heterogeneous aircraft styles and variation in sizes. Top Row:} Small, Medium and Large Civil Transports. \textbf{Bottom Row:} Military Fighter, Bomber, and Transport. }
  \label{roles}
\end{center}
\end{figure}
\end{itemize}

\begin{figure*}[!htb]
\centering
  \includegraphics[width=0.8\textwidth]{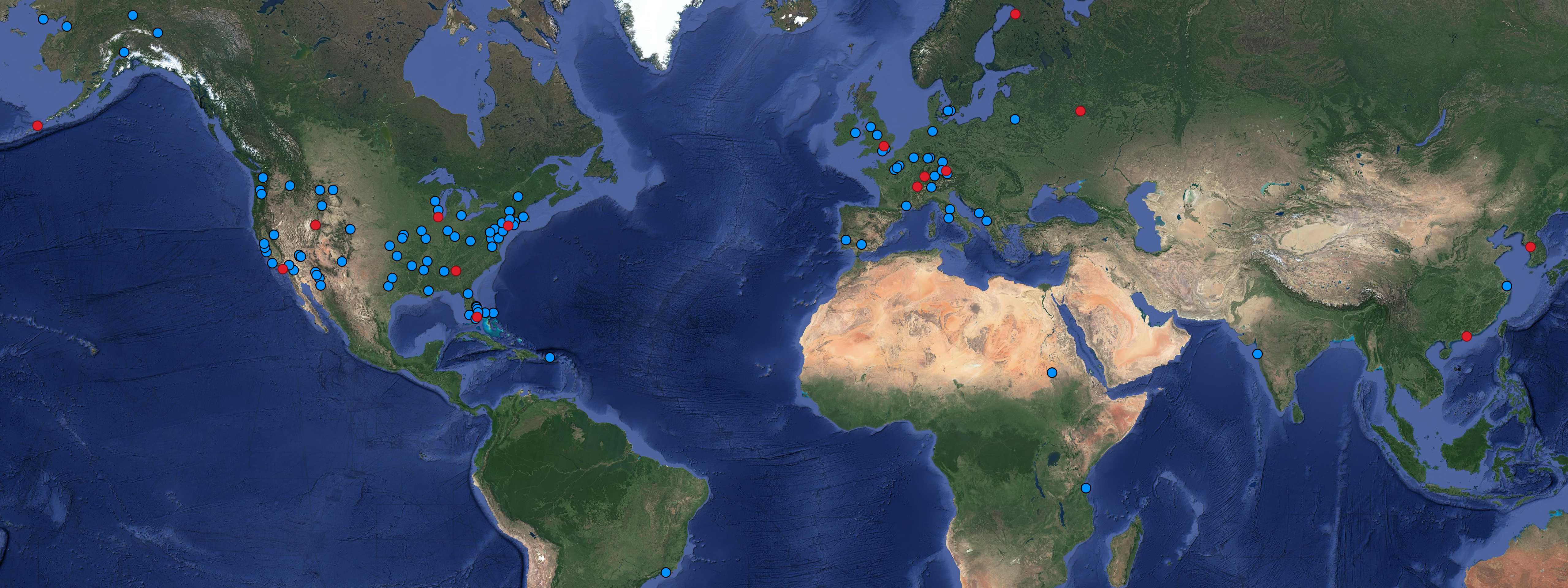}\caption{\textbf{RarePlanes dataset locations.}  The dataset features 112 real (blue points) and 15 synthetic locations (red points).  Atlanta, Miami, and Salt Lake City feature both real and synthetic data.}
  \label{Map}
  \vspace{-1mm}
\end{figure*}

\subsection{Real Imagery and Locations}

All electro-optical imagery is provided by the Maxar Worldview-3 satellite with a maximum ground sample distance (GSD) of $0.31$ to $0.39$ meters depending upon sensor look-angle. The dataset consists of $253$ unique scenes, spanning $2,142 \, \rm{km}^2$ with $112$ locations in $22$ countries.  Locations were chosen by performing a stratified random sampling of OpenStreetMap \cite{osm} aerodromes of area $\geq 1 \, \rm{km}^2$ across the US and Europe using the K{\"o}ppen climate zone as the stratification layer. We stratify by climate to increase seasonal diversity and geographic heterogeneity. Seven additional locations were manually chosen as they overlap with preexisting datasets \cite{weir2019spacenet,etten2018spacenet}, and we considered further revisits over these locations to potentially have additional value. We then chose individual satellite scenes by attempting to select scenes from different seasons for each location. Many locations have several scenes taken at different points in time, which may enable future investigation on the value of annotating the same areas using multiple images. The imagery is collected from variable look-angles ($3.2$ to $29.6^{\circ}$), target azimuth angles ($1.8$ to $359.7^{\circ}$), and sun elevation angles ($10.7$ to $79.0^{\circ}$). Imagery is collected from all four seasons, with scenes featuring instances of cloud cover ($12.6\%$), snow ($9.1\%$) and clear skies ($78.3\%$).  Combined together, this leads to high variability in illumination, shadowing, and lighting conditions. Consequently, the dataset should help to improve generalizability to new areas.  Finally, background surfaces are quite diverse with grass, dirt, concrete, and asphalt surface types.

The collection is composed of three different sets of data with different spatial resolutions: one panchromatic band ($0.31-0.39$m), eight multi-spectral (coastal to NIR ($400-954 \mu$m)) bands ($1.24-1.56$m), and three RGB ($448-692 \mu$m) pan-sharpened bands ($0.31-0.39$m). Each data product is atmospherically compensated to surface-reflectance values by Maxar's AComp \cite{AComp} and ortho-rectified using the SRTM DEM.  RGB data is also converted to 8-bit. Areas containing non-valid imagery are set to $0$.  We distribute both $512 \times 512$ pixel tiles ($20\%$ overlap) that contain aircraft as well as as the full images, cropped to the extent of the area of annotation.

\subsection{Synthetic Data}


The synthetic dataset contains $629,551$ annotations of aircraft across $50,000$ images with an extent of $1920 \times 1080$ pixels at 15 distinct locations,  simulating a total area of $9331.2\, \rm{km}^2$. All synthetic data is created via AI.Reverie’s novel simulator software in a multi-step process. First, individual real-world locations are simulated using a proprietary procedural GIS framework integrated with Houdini \cite{houdini}. The framework ingests geospatial vector data (OpenStreetMap \cite{osm}) and overhead images to procedurally generate building and airport terminal models. Airport ground and runways are then generated through simple planar shape and texture projections.  Once the simulated locations are generated, we import each location into unreal engine 4 \cite{unreal}.  Capture modules and settings are configured to spawn aircraft and simulate the environmental factors and biomes.  Each image features a simulated GSD of $0.3$ meters to closely approximate our real imagery and is collected from variable look-angles ranging between $5.0$ to $30.0^{\circ}$ off-nadir.  The imagery is evenly split across 5 distinct biomes including: `Alpine', `Arctic', `Temperate Evergreen Forests', `Grasslands', and `Tundra'. The biome parameter controls the type of vegetation, its density, as well as the ground textures. Four unique weather conditions are also evenly distributed across the dataset including: `Overcast', `Clear Sky', `Snow', and `Rain'. Other parameters include the sunlight intensity, weather intensity, and the time of the day. Ultimately, this produces an expansive heterogeneous dataset with a wide variety of backgrounds. We believe that this dataset will be helpful in improving model generalizability to new areas and developing new algorithmic approaches that could move beyond aircraft detection.

\section{Experiments, Results, and Discussion}

\begin{table*}[!htb]
  \centering
  \begin{tabular}{lcccccccc}
    \toprule
    network & attribute & dataset & $C_{S}$ & $C_{M}$ & $C_{L}$ & mAP & mAP50 & AR \\
    \midrule
     & aircraft & Real & N/A & N/A & N/A & 73.32 (0.34) & 96.80 (0.02) &  77.16 (0.21)\\
     & aircraft & Synth. & N/A & N/A & N/A & 54.86 (0.25) & 87.03 (0.53) &  60.67 (0.27)\\
    \multirow{2.5}{3.5em}{Faster R-CNN}  & aircraft & FT & N/A & N/A & N/A & 69.16 (0.69) & 95.29 (0.41) & 73.03 (0.57)  \\
    \cmidrule(r){2-9}
     & role & Real & 66.68 & 70.26 & 67.68 & 68.21 (0.4) & 92.16 (0.23) & 75.39 (0.40) \\
     & role & Synth. & 27.70 & 37.09 & 42.85  & 35.88 (2.26) & 59.09 (2.9) & 53.82 (1.28)\\
     & role & FT & 56.73 & 66.05 & 66.52 & 63.10 (0.78) & 89.15 (0.22) & 71.06 (0.75)\\
    \midrule
     & aircraft & Real & N/A & N/A & N/A & 73.67 (0.17) & 96.81 (0.03) & 76.46 (0.20)\\
     & aircraft & Synth. & N/A & N/A & N/A & 56.28 (0.46) & 87.54 (0.69) & 60.71 (0.51)\\
    \multirow{2.5}{3.5em}{Mask R-CNN} & aircraft & FT & N/A & N/A & N/A & 70.51 (0.34) & 94.73 (0.03) & 73.72 (0.26) \\
    \cmidrule(r){2-9}
     & role & Real & 65.60 & 72.13 & 70.97 & 69.57 (0.47) & 91.89 (0.55) & 76.16 (0.30)\\
     & role & Synth. & 29.12 & 41.78 & 47.47 & 39.46 (3.20) & 62.31 (4.51) & 57.33 (1.96)\\
     & role & FT & 58.96 & 70.02 & 72.33 & 67.11 (0.46) & 90.03 (0.52) & 74.40 (0.58)\\
    \bottomrule
  \end{tabular}
    \vspace{3mm}
    \caption{\textbf{Results of the object detection and segmentation experiments.} We report models performance trained on the real dataset (Real) and the synthetic dataset (Synth.) as well as the fine tuning experiment (FT) using only $10\%$ of the real training dataset. We show the results of the single class experiments (`aircraft') and the three classes experiment: small ($C_{S}$), medium ($C_{M}$), and large ($C_{L}$) civil transport aircraft. Performance is evaluated using the mean average precision (mAP) (IOU@[0.5:0.95]), the mAP50 (IOU@0.5) and the average recall (AR) metrics, as well as the class APs when applicable. For the Mask R-CNN instance segmentation experiments, we only report the segmentation AP. Each value reported is an average of 5 runs. The standard deviations for mAP, mAP50, and AR  are also indicated.}
  \label{results-table}
\end{table*}

In this section, we validate the synthetic dataset by running three experiments for two tasks: object detection and instance segmentation. For each task, we train a benchmark network on three subsets of data: on the real data only, on the synthetic data only, and perform a fine tuning experiment training on the synthetic data and then a portion ($\sim10\%$) of the real dataset. Each experiment is validated on the test real dataset and the results are shown Table \ref{results-table}. We ran these experiments for two attributes: aircraft (detection of an aircraft without classifying it) and civil role.

\subsection{Training and Testing Splits}
For the real world data, given the size of the raw satellite scenes, we adopted a tiling approach. Each scene is split into 512x512 tiles containing at least one aircraft. Furthermore, we ensure that the training and test split contains at least one satellite scene per country to maximize geographic diversity. As the dataset contains satellite images captured over the same location at different dates, an airfield can appear in both splits but at different points in time. Moreover, we created a subset of the real training split for the fine tuning experiments. This subset contains roughly 10\% of the images of the training split, created by drawing a 10\% random sample of image tiles by location. All synthetic data is included in the training split to maintain a domain gap. For this study we train on $45,000$ of the $50,000$ images, reserving $5,000$ images for cross-validation purposes. Note these cross-validation results are not reported here. Further details on the initial training and testing splits used in this study can be found at \url{https://www.cosmiqworks.org/RarePlanes}. Ultimately end users are encouraged to reorganize and create their own splits for their specific research or use-case(s).

\begin{figure*}[!htb]
\setlength{\tabcolsep}{0.2em}
\begin{center}
  \begin{tabular}{cccc}
    \includegraphics[width=0.2\textwidth]{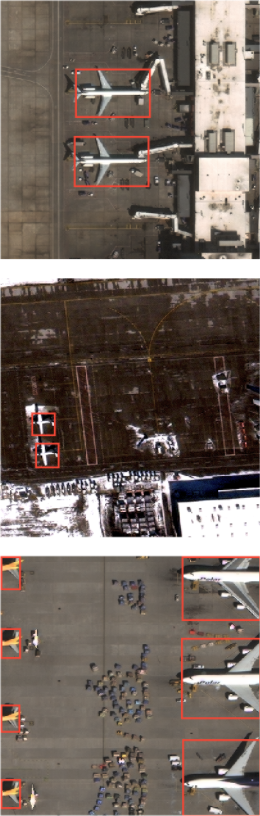} & \includegraphics[width=0.2\textwidth]{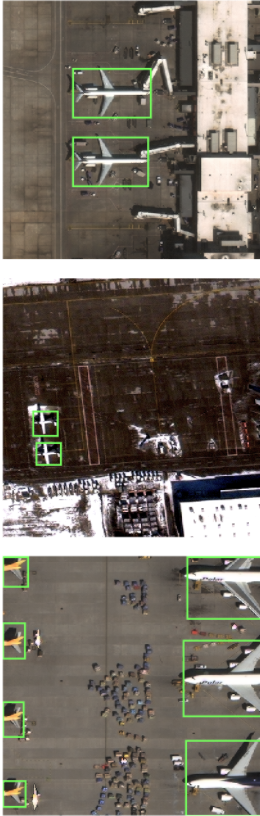} & \includegraphics[width=0.2\textwidth]{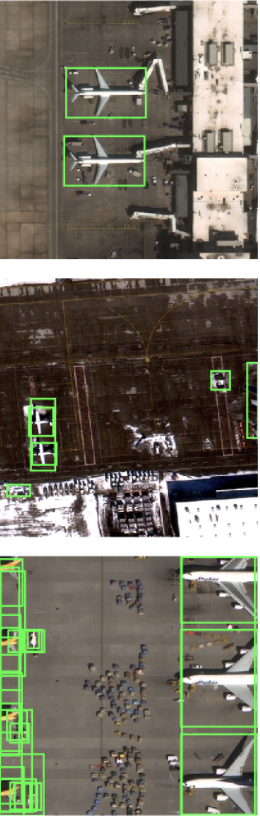} & \includegraphics[width=0.2\textwidth]{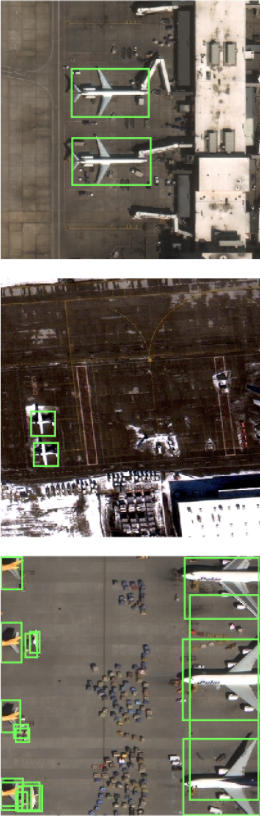} \\
    (a) & (b) & (c) & (d) \\
  \end{tabular}
  \end{center}
  \caption{\textbf{Example of aircraft detection results}. (a) ground truth, (b) model trained real dataset (c) model trained on synthetic dataset (d) model fine tuned on real subset.}
  \label{detection-examples}
    \vspace{-2mm}
\end{figure*}

\subsection{Implementation}
In our experiments, we used a Resnet-50 \cite{he2015deep} and FPN \cite{lin2016feature} as the backbone for the Faster R-CNN \cite{ren2015faster} detection network. A similar backbone was used for the Mask R-CNN \cite{he2017mask} instance segmentation network. Backbones are  pre-trained using ImageNet \cite{deng2009imagenet}weights and all experiments are conducted with the Detectron2 framework \cite{wu2019detectron2}, using the default configurations for each network.  The network was optimized with Stochastic Gradient Descent (SGD) using a learning rate of 0.001, weight decay of 0.0001 and a momentum of 0.9. Additionally, we used a linear warmup period over 1K iterations. We maintain a consistent learning rate for the fine tuning experiments. We found that decreasing the learning rate or freezing some of the layers in the backbone did not improve performance. The networks were trained on a NVIDIA Tesla V100 GPU with 12GB memory. Each network was trained until convergence, which was reached after around 60K iterations. We also applied basic pixel level augmentations, such as blurring and modifying the contrast or the brightness. Finally, we performed random cropping (512x512) when training on the synthetic dataset.

\subsection{Results and Discussion}
We evaluated our network performances using the COCO average precision (AP) metric. Table \ref{results-table} reports the average precision for each class as well as the mAP, mAP50, and average recall (AR). Qualitative results are shown in Figure \ref{detection-examples}.

In the first set of experiments, we focused on the performance of the synthetic dataset only. As expected, we observe a drop in performances when training on the synthetic data only, due to the domain gap between the real and synthetic datasets. We observe that the model trained on the synthetic dataset tends to mislabel clutter or nearby objects as aircraft, as shown in Figure \ref{detection-examples}. Additionally, snow patches, ground markings, airport vehicles are sometimes detected as aircraft. This leads to a significantly lower AP ($55\%$ to $75\%$ of the real AP)  when models are trained on the synthetic dataset only. However, the AR is not as sensitive to the domain gap ($70\%$ to $80\%$ of the real AR), meaning that the majority of aircraft are still detected when only the synthetic dataset is used. Similarly, we observe that the drop in AP50 is also lower relative to the AP metric. Ultimately, the AP50 metric may be more informative as we are most interested in accurately counting aircraft, rather than how well they are localized.

Most importantly, when a small subset ($\sim10\%$) of real data is added for fine tuning, we observe a significant gain in mAP, leading to similar performance (91 to $96\%$ of real mAP) to the models trained on the real dataset only. We hypothesize that the synthetic data helps to build a prior model for aircraft detection and eases transfer learning, thus greatly reducing the need for annotated real data. In Figure \ref{detection-examples}, we see how fine tuning on the real subset removes some of the false positive predictions versus training on the synthetic dataset only. However, the false positive detection rate still remains slightly higher compared to training on the entire real training set. It's important to note that the goal of these experiments is to define a baseline for future experimentation for other algorithms to improve upon, particularly within the area of domain adaptation. 

\section{Conclusions}

We present RarePlanes, a unique machine learning dataset that incorporates both real and synthetically generated satellite imagery. This dataset is critical for expanding and examining the value of synthetic data for overhead applications.  Our benchmark experiments using the dataset found that blends of $90\%$ synthetic to $10\%$ real can deliver nearly equivalent performance as $100\%$ real data for the task of aircraft identification. Additionally, we believe that RarePlanes could be particularly valuable for developing new and extensible domain adaptation approaches. Finally, the detailed fine-grain attribution and annotation contained in the dataset may enable various CV tasks and future experiments in the fields of detection, instance segmentation, or zero-shot learning.

\section*{Acknowledgments} 
The authors thank the whole AI.Reverie team for making the creation of the synthetic dataset possible. We would especially like to thank Danny Gillies and Natasha Ruiz for their devoted help.  Additionally we thank Christyn Zehnder at IQT for her contributions in launching and marketing the RarePlanes dataset.


{\small
\bibliographystyle{ieee_fullname.bst}
\bibliography{RarePlanes_cites.bib}
}

\end{document}